\documentclass{article}

\usepackage[english]{babel}

\usepackage[letterpaper,top=2cm,bottom=2cm,left=3cm,right=3cm,marginparwidth=1.75cm]{geometry}

\usepackage{amsmath}
\usepackage{graphicx}
\usepackage[colorlinks=true, allcolors=blue]{hyperref}
\usepackage{booktabs}
\usepackage{multirow}
\usepackage[english]{babel}

\usepackage[capitalize]{cleveref}
\crefname{section}{Sec.}{Secs.}
\Crefname{section}{Section}{Sections}
\Crefname{table}{Table}{Tables}
\crefname{table}{Tab.}{Tabs.}

\title{Feature Diversity Learning with Sample Dropout for Unsupervised Domain Adaptive Person Re-identification}
\author{Chunren Tang,Dingyu Xue,Dongyue Chen}

\begin{document}
\maketitle

\begin{abstract}
Clustering-based approach has proved effective in dealing with unsupervised domain adaptive person re-identification (ReID) tasks. However, existing works along this approach still suffer from noisy pseudo labels and the unreliable generalization ability during the whole training process. To solve these problems, this paper proposes a new approach to learn the feature representation with better generalization ability through limiting noisy pseudo labels. At first, we propose a Sample Dropout (SD) method to prevent the training of the model from falling into the vicious circle caused by samples that are frequently assigned with noisy pseudo labels. In addition, we put forward a brand-new method referred as to Feature Diversity Learning (FDL) under the classic mutual-teaching architecture, which can significantly improve the generalization ability of the feature representation on the target domain. Experimental results show that our proposed FDL-SD achieves the state-of-the-art performance on multiple benchmark datasets.
\end{abstract}

\section{Introduction}
\label{sec:intro}

Person re-identification (ReID) aims to match person images across multiple non-overlapping cameras, which has achieved attention from both industry and academia. Most of the existing person ReID models along the supervised approach \cite{ref1:1,ref1:2,ref1:3,ref1:4,ref1:5} have achieved satisfactory performance. However, these models generally perform less well in real applications because they have never been trained to adapt to the application scenes. To address this issue, a new problem referred as to unsupervised domain adaptation becomes a hot topic in ReID task \cite{ref:ECN,ref1:6,ref1:7}, which focuses on how to adapt a pretrained model from a labeled source domain to an unlabeled target domain.

The main challenge of unsupervised domain adaptive person ReID lies in learning feature representation with unlabeled target domain data. To solve this challenge, one major line attempts to assign pseudo labels for target samples based on the pretrained model trained with labeled source samples \cite{ref:ECN,ref:SpCL,ref:DCML,ref:NRMT}, and then fine-tunes the model using the target samples with pseudo labels. Obviously, following this approach, the person ReID performance highly depends on the quality of pseudo labels. Therefore some works focus on obtaining highly dependable pseudo labels. Some of these works concentrate on the clustering process \cite{ref:SpCL,ref:DCML,ref:JVTC}, in which target samples are assigned with pseudo labels based on different metrics. This process can improve the clustering result. Other works aim at how to effectively utilize target samples based on the clustering results \cite{ref:NRMT,ref:UNRN}. During the training process of these methods, the pseudo labels with different reliability are assigned with different weights. However, due to the clustering results are unsatisfactory, all of these methods suffer from noisy pseudo labels. Experimental results revealed that a small proportion of the samples are assigned with wrong pseudo labels frequently. These samples can be regarded as hard samples for the unsupervised domain adaptive person ReID task. As is well known, the performance of a well-trained model relies more on hard samples than easy samples. For the same reason, hard samples with stubborn wrong pseudo labels will limit the performance of the ReID model heavily. Unfortunately, existing unsupervised domain adaptive ReID methods can hardly solve this problem. Meanwhile, most of these models only apply supervised loss functions such as the Cross Entropy and Triplet loss based on the pseudo labels to train the model, but neglect unsupervised feature learning without ground-truth labels or pseudo labels, which hiddenly limits the generalization ability of the learned feature representation.

To address these two problems, we propose a novel solution FDL-SD to resist hard samples and learn features with better generalization ability in a united framework. In order to limit the ill influence of these hard samples, we propose a simple but powerful method which is called Sample Dropout (SD) to smooth the distribution of noisy pseudo labels. For most clustering-based unsupervised domain adaptive ReID works, assigning pseudo labels for all target samples is employed before each fine-tuning iterator. But in this paper, we randomly discard a proportion of target samples before each epoch of the training, through which the vicious circle of iterative training caused by hard samples can be broken. In addition to the proposed SD, we also present a new architecture to realize Feature Diversity Learning (FDL) in an unsupervised way, which is believed to suppress the ill effect of wrong pseudo labels and enhance the generalization ability of the feature representation. Overall, the main contributions of this paper can be summarized in three aspects:

(1) We propose the Sample Dropout (SD) method to reduce the adverse effect of hard samples on domain adaptation, which can prevent some hard samples from being assigned with wrong pseudo labels all the time, thus breaking the vicious circle caused by these hard samples.

(2) We propose the Feature Diversity Learning (FDL) and embed it into a dual-branch architecture to learn feature diversity representation in an unsupervised fashion, which can boost the generalization ability of model on target domain.

(3) Extensive experiments on multiple benchmark datasets show that our proposed FDL-SD  achieves the state-of-the-art performance, which demonstrates the effectiveness of our proposed approach.

\section{Related Work}
\label{sec:related}
\subsection{Unsupervised Domain Adaptive Person ReID}
Unsupervised domain adaptive person ReID aims at transferring the knowledge learned from a labeled source domain to an unlabeled target domain. Existing works can be roughly divided into two categories. For the first category, it attempts to reduce the domain gap between the labeled source domain and unlabeled target domain. Some methods reduce the discrepancy between two domains by aligning the feature distribution\cite{ref2:1,ref:EaNet,ref:DMMD}. Some other methods adopt Generative Adversarial Network (GAN) technology to transfer the person images from source style to target style\cite{ref:PTGAN,ref:SPGAN,ref:CRGAN,ref:DGNET}. However, the identify information of the target domain is ignored in these methods, where the target samples are not fully utilized to train the model. For the second category, it assigns pseudo labels for target samples based on the pretrained model trained with labeled source domain data, and then fine-tunes the model in supervised fashion \cite{ref:BUG,ref:HCT,ref:SpCL,ref:MMT,ref:NRMT,ref:MEBNet}. These methods are widely utilized because of their superior performance. BUC\cite{ref:BUG}, HCT\cite{ref:HCT} and SpCL\cite{ref:SpCL} directly fine-tune the model relying on the iteration of pseudo-label mining. Some other methods adopt mutual-training to cluster target samples and train the model, such as MMT\cite{ref:MMT}, MEB-Net\cite{ref:MEBNet} and NRMT\cite{ref:NRMT}. But the wrong pseudo labels are inevitable in these approaches, and some hard samples may even seriously damage the model.

In this paper, we adopt the second line to solve the unsupervised domain adaptive person ReID task, but the difference compared with the existing works is that our work is motivated to reduce the ill effect caused by hard samples and explore feature diversity representation. 

\subsection{Learning with Noisy Labels}
Some remarkable methods have been devoted to handling noisy labels, which can be categorized into loss correction, sample reweighting and label correction. Loss correction methods engage to design special loss functions against noisy labels \cite{ref:MAE,ref:SCE,ref:GCE}. Sample reweighting methods assign various weights to different samples \cite{ref:UNRN,ref:meta-weight,ref:learningwith,ref:Co}. For example, UNRN\cite{ref:UNRN} proposed to re-weight samples based on the uncertainty of their pseudo labels. Label correction methods \cite{ref:DCML,ref:twostagecluster,ref:JVTC,ref:cleannet,ref:towards} focus on direct correction of noisy labels. DCML\cite{ref:DCML} proposed to gradually increase the usage of pseudo labels as the training process goes on, which helps the model learn knowledge from easy to hard. ADTC\cite{ref:twostagecluster} proposed to use a two-stage clustering strategy to assign pseudo labels for target samples, which can improve the clustering quality. However, all of these methods can’t solve the problem that some hard samples with noisy labels hurt the training of the model seriously and are difficult to detect or correct in the iterative optimization process. Therefore, this paper adopts a novel Sample Dropout method to weaken the ill influence caused by theses hard samples.

\subsection{Unsupervised Feature Learning}
There are no ground-truth labels in some classification tasks, therefore the unsupervised methods are adopted. MMCL\cite{ref:MMCL} designed a memory-based multi-label classification loss which integrates multi-label classification and single-label classification in a unified framework. Similar to MMCL, Xiao et al \cite{ref:OIM} introduced a parameter-free Online Instance Matching loss with a memory dictionary scheme, which trains feature encoder directly instead of needing to learn a big classifier matrix. In order to mitigate the effects of noisy pseudo labels, MMT\cite{ref:MMT} introduced a novel soft softmax-triplet loss to support learning with soft pseudo labels. To solve the problem that previous contrastive losses\cite{ref:CL1,ref:CL2,ref:CL3} only focused on separating instances without considering any ground-truth classes or pseudo-class labels, SpCL\cite{ref:SpCL} proposed a unified contrastive loss jointly 
distinguishes source domain classes, clusters and un-clustered instances of target domain. However, all of these methods need to adopt pseudo labels as supervised signals, but the wrong pseudo labels are inevitable. In this paper, we propose the Feature Diversity Learning (FDL), which does not need labels as supervision that avoids the noisy labels affecting the model training. At the same time, it can boost the model’s representation ability.

\section{Method}
\label{sec:method}
\subsection{Overall Framework}
\begin{figure*}
  \centering
  \includegraphics[width=0.8\textwidth]{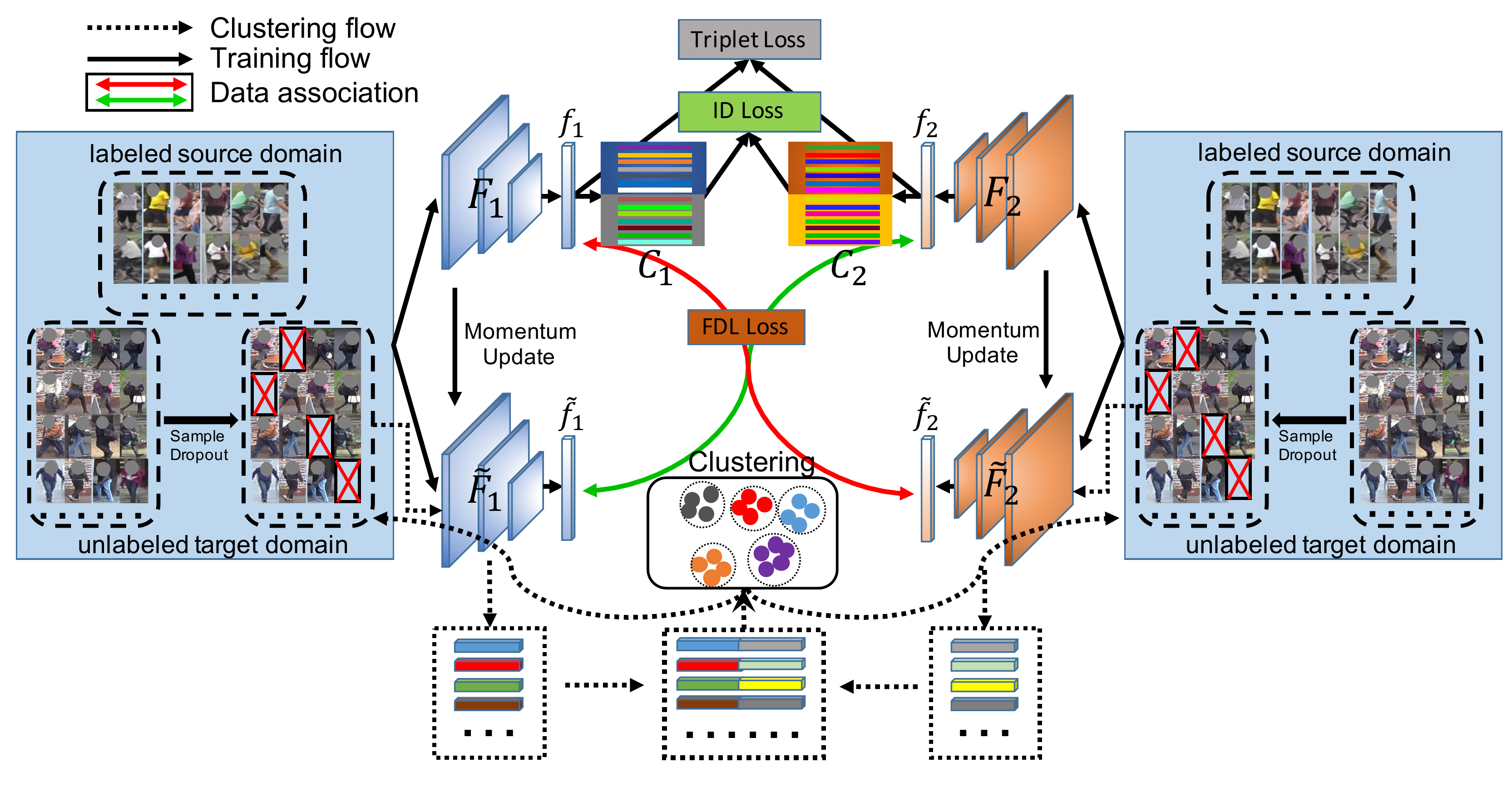}
  \caption{An overview of the proposed architecture. Our framework adopts a dual-branch structure which is consisted of feature encoders $F_1$, $F_2$ and their corresponding mean feature encodes $\tilde{F}_1$ , $\tilde{F}_2$. Classifiers $C_1$ and $C_2$ follow behind $F_1$ and $F_2$.}
  \label{fig:framework}
\end{figure*}

For the unsupervised domain adaptive person ReID task, we have a labeled training dataset $D^{\rm s}=\left\{(\textbf{\emph{x}}_i^{\rm s},y_i^{\rm s})\right\}_{i=1}^{N_{\rm s}}$ collected from the source domain, where $\textbf{\emph{x}}_i^{\rm s}$ and $y_i^{\rm s}$ denote the $i$-th source sample and its corresponding person identity label, $N_{\rm s}$ is the number of all the source samples for training. The unlabeled target domain samples are denoted as $D^{\rm t}=\left\{{\textbf{\emph{x}}_i^{\rm t} }\right\}_{i=1}^{N_{\rm t}}$. Before each iterative training epoch, the clustering algorithm DBSCAN\cite{ref:DBSCAN} will be used to assign pseudo labels for target samples. In order to reduce the ill influence of hard samples in the target domain and boost the generalization ability of the model, we propose a novel framework that contains Sample Dropout and Feature Diversity Learning, as shown as \cref{fig:framework}.  

Our model adopts the dual-branch structure which consists of feature encoders $F_1$ and $F_2$. Correspondingly, we leverage momentum update mechanism to construct and update two mean feature encoders $\tilde{F}_1$ and $\tilde{F}_2$ respectively. During the training process, two mini-batches are randomly selected from source domain and target domain, and they are fed into the two branches. A point to note is that the proposed SD step has been applied in the target domain before each training epoch. Along each branch, the output feature vector  $\textbf{\emph{f}}_i$ of the encoder $F_i$ will be sent into the classifier $C_i$.

As shown in \cref{fig:framework}, the classic Cross Entropy loss is deployed based on the ID predictions given by the classifier $C_i$, the Triplet loss is calculated based on the feature vector $\textbf{\emph{f}}_i$, and the proposed FDL loss is embedded into the mutual teaching architecture based on the encoder’s output  $\textbf{\emph{f}}_i$ and mean feature encoder’s output $\boldsymbol{\tilde{f}}_j,i\neq j$. At last, the feature vectors of two mean encoders will be concatenated into a united vector $\boldsymbol{\tilde{f}}=[\boldsymbol{\tilde{f}}_1;\boldsymbol{\tilde{f}}_2]$ to represent each sample in the testing stage, which will be also used by the clustering algorithm to produce pseudo labels.

In summary, there are three main steps during each epoch of the iterative training: 1) Sample Dropout in the target domain, 2) assigning pseudo labels for target domain samples, 3) training the model with source samples and target samples. Through the iterative optimization between feature representations and pseudo labels, the performance of the proposed method will be effectively improved. 

\subsection{Sample Dropout}
Noisy pseudo labels are harmful to the model but inevitable for unsupervised domain adaptive ReID tasks. However, we found that a small part of samples often account for a large proportion of noisy pseudo labels during the whole training process, which we define as hard samples. Compared with the general samples which are assigned with wrong pseudo labels only a few times during the whole training process, hard samples will continuously mislead the training process to wrong directions and irreversibly damage the model. In order to solve this problem, we propose Sample Dropout to smooth the distribution of noisy pseudo labels, which can reduce the adverse effect caused by hard samples.

Sample Dropout is adopted before the clustering step of each iterative training epoch. In the beginning of $k$-th iterative epoch, a proportion of samples are randomly selected from the target dataset and denoted as $D_k^{\rm t}=\left\{\textbf{\emph{x}}_{r(j,k)}^{\rm t}\right\}_{j=1}^{M_{\rm t}}$, where $M_{\rm t}=(1-\rho)N_{\rm t}$, $\rho$ represents the Sample Dropout rate. Function $r(j,k)$ indicates the $j$-th random sample from the target dataset $D^{\rm t}$ in the $k$-th epoch. Then only the selected samples $D_k^{\rm t}$ will be assigned with pseudo labels based on their clustering results, and the residual target samples will be dropped out from the current training epoch.

\begin{figure}[t]
  \centering
   \includegraphics[width=0.45\textwidth]{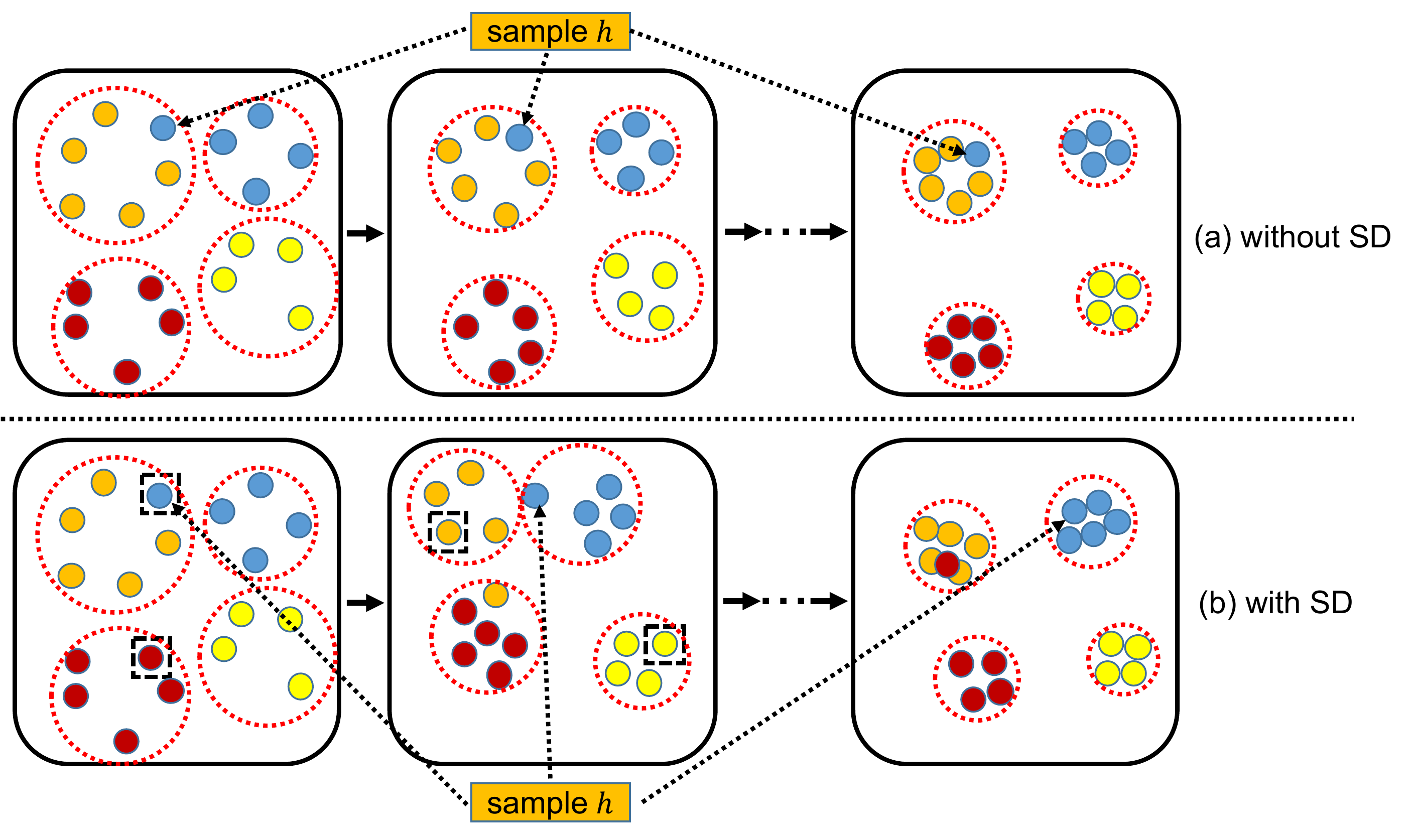}
   \caption{Illustration of the clustering results before each fine-tuning epoch. Circles with the same color represent samples with the same person ID. Samples in the same red dotted circle belong to the same cluster. The upper row (a) represents a typical clustering process without SD, and the bottom row (b) represents the process with SD, in which the black dotted boxes mark the discarded samples in SD step.}
   \label{fig:SD}
\end{figure}

Though the proposed SD seems extremely simple and naïve, it proves very powerful in dealing with noisy pseudo labels because it can prevent the model from accumulating degradation caused by hard samples. As is shown in \cref{fig:SD}, in upper row, the sample $h$ is a hard sample which is always assigned with wrong pseudo labels. As mentioned above, such a hard sample with wrong labels can easily lead to the vicious circle of iterative training between wrong label and bad feature so that the feature space will be forced to gather samples with different person IDs while scattering the samples with the same ID, just as shown as the top row of \cref{fig:SD}. Considering that the query in a traditional ReID task is always based on the feature distance between two samples, such a distorted feature space will noticeably degrade the performance of the model. In contrast, as shown as \cref{fig:SD}(b), the employment of SD can occasionally stop these hard samples from participating in the clustering and training steps and consequently prevent the training process from falling into the local minimum trap caused by these hard samples. Without the disturbance of hard samples, the feature space can be pulled back from the wrong direction to a better direction in the next epoch. As a result, there is larger probability to assign these hard samples with correct pseudo labels, as shown as the second plot in the bottom row of \cref{fig:SD}. By repeatedly implementing SD, the negative impact of hard samples on the whole training process can be effectively suppressed, and the final performance of the model will be dramatically improved.

\subsection{Feature Diversity Learning}
Most unsupervised domain adaptive ReID methods employ only pseudo labels to guide the training of the model. On one hand, noisy pseudo labels could hurt the model severely. On the other hand, the generalization ability of the feature representation has not been sufficiently boosted because only the pseudo-label-based losses are applied in the training stage. Therefore, we present a new approach referred as to Feature Diversity Learning (FDL), in which the pseudo labels are not involved, and the generalization ability can be effectively enhanced.  

Our model adopts a dual-branch structure to produce two feature streams which will be concatenated together at the final stage to form a united feature representation. The diversity between the two streams can be regarded as a kind of regularization term. It is believed to benefit the generalization ability of the feature representation. The main idea of FDL is about how to make both the two streams serve to the same ReID task while keeping their diversity to each other. The biggest challenge lies on the trade-off of the similarity and the diversity between the two feature streams. Higher similarity helps to speed up the convergence and reduce the empirical error but take higher risk of overfitting. In contrast, keeping appropriate feature diversity generally indicates better generalization ability but makes the training difficult to converge. To get a stable and proper balance, as shown as \cref{fig:framework}, we build a mutual teaching architecture between the two streams by constructing a mean feature encoder $\tilde{F}_i$ for each stream and updating it with the momentum of the corresponding feature encoder, which can be demonstrated as \cref{eq:1}
\begin{equation}
  \boldsymbol{\tilde{{\theta}}}_i(T)=\alpha\boldsymbol{\tilde{\theta}}_i(T-1)+(1-\alpha)\boldsymbol{\theta_i}(T)
  \label{eq:1}
\end{equation}
where $\boldsymbol{\theta}_i$ and $\boldsymbol{\tilde{\theta}}_i$ are the parameters of $F_i$ and $\tilde{F}_i$ respectively, $T$ and $(T-1)$ represent the current statement and previous iterative statement, and $\alpha$ is the momentum coefficient. In addition, the FDL loss function $\mathcal{L}_{\rm FDL}$ is put forward to guide the feature diversity learning within the proposed architecture, as shown in \cref{eq:2}.

\begin{equation}
  \mathcal{L}_{\rm FDL}=S(\textbf{\emph{f}}_1^\mathrm{T}\boldsymbol{\tilde{f}}_2)+S(\textbf{\emph{f}}_2^\mathrm{T}\boldsymbol{\tilde{f}}_1)
  \label{eq:2}
\end{equation}
where $S(x)$ is a soft-plus function.

\begin{equation}
  S(x)={\rm ln}(1+{\rm exp}(x))
  \label{eq:3}
\end{equation}

It is easy to understand that the more diverse the vectors $\textbf{\emph{f}}_i$ and $\boldsymbol{\tilde{f}}_j$, $i\neq j$ are, the smaller the FDL loss $\mathcal{L}_{\rm FDL}$ is. With the proposed mutual teaching structure embedded with the FDL loss, stable and proper diversity between the two branches can be expected. Experimental results prove that the proposed Feature Diversity Learning is effective. We believe this is because the two branches will converge to different suboptimal solutions under the guidance of the FDL loss, which helps lower the risk of overfitting to the outlier samples or unreliable pseudo labels and therefore enhances the generalization ability of the feature representation. 

\subsection{Overall Loss}
In addition to the FDL loss, we also adopt the widely used Cross Entropy (CE) loss and the Triplet loss to train the model. 

The CE loss is assigned based on the predictions given by both the classifiers $C_1$ and $C_2$. For the source mini-batch $B^{\rm s}$ and target mini-batch $B^{\rm t}$, the CE loss $\mathcal{L}_{\rm CE}$  is defined as follows:

\begin{equation}
\begin{aligned}
  \mathcal{L}_{\rm CE}=-[&\frac{1}{|B^{\rm s}|}\sum_{\textbf{\emph{x}}_i\in{B^{\rm s}}}({\rm log}(\hat{y}_{i,*}^{(1)})+{\rm log}(\hat{y}_{i,*}^{(2)}))+\\
  &\frac{1}{|B^{\rm t}|}\sum_{\textbf{\emph{x}}_j\in{B^{\rm t}}}({\rm log}(\hat{y}_{j,\#}^{(1)})+{\rm log}(\hat{y}_{j,\#}^{(2)}))]
  \label{eq:4}
 \end{aligned}
\end{equation}
where $\hat{y}_{i,*}^{(1)}$ and $\hat{y}_{i,*}^{(2)}$ denote the predicted probabilities of the source sample $\textbf{\emph{x}}_i^{\rm s}$ on its true label, which are obtained by the classifiers $C_1$ and $C_2$ respectively. Similarly, $\hat{y}_{j,\#}^{(1)}$ and $\hat{y}_{j,\#}^{(2)}$ are those of the target sample $\textbf{\emph{x}}_j^{\rm t}$ on the pseudo label.

The standard Triplet loss for a training sample $\textbf{\emph{x}}_a$ (the anchor sample) and its feature vector $F(\textbf{\emph{x}}_a)$ can be calculated according to \cref{eq:5}.
\begin{equation}
\begin{aligned}
  \mathcal{L}_{\rm t}(\textbf{\emph{x}}_a;F)=[\tau+&\left\|F(\textbf{\emph{x}}_a)-F(\textbf{\emph{x}}_p))\right\|^2 - \\ &\left\|F(\textbf{\emph{x}}_a)-F(\textbf{\emph{x}}_n))\right\|^2 ]_{+}
  \label{eq:5}
 \end{aligned}
\end{equation}
where $\textbf{\emph{x}}_p$ and $\textbf{\emph{x}}_n$ are the hardest positive and negative samples for the anchor $\textbf{\emph{x}}_a$ in the same mini-batch. By applying the Triplet loss $\mathcal{L}_{\rm t}$ to both the source and target mini-batches in two branches, the overall Triplet loss can be written as \cref{eq:6}. It should be noted especially that the hardest positive and negative samples of a target sample are selected according to their pseudo labels.

\begin{equation}
\begin{aligned}
  \mathcal{L}_{\rm TRI}=&\frac{1}{|B^{\rm s}|}\sum_{\textbf{\emph{x}}_i\in{B^{\rm s}}}(\mathcal{L}_{\rm t}(\textbf{\emph{x}}_i;F_1) + \mathcal{L}_{\rm t}(\textbf{\emph{x}}_i;F_2) )+ \\
  &\frac{1}{|B^{\rm t}|}\sum_{\textbf{\emph{x}}_j\in{B^{\rm t}}}(\mathcal{L}_{\rm t}(\textbf{\emph{x}}_j;F_1) + \mathcal{L}_{\rm t}(\textbf{\emph{x}}_j;F_2) )
  \label{eq:6}
 \end{aligned}
\end{equation}

At last, the CE loss, Triplet loss and the proposed FDL loss are combined together to train the model in an end-to-end fashion. The overall loss function can be written as:

\begin{equation}
\begin{aligned}
  \mathcal{L}=\beta\mathcal{L}_{\rm CE}+\gamma\mathcal{L}_{\rm TRI}+\delta\mathcal{L}_{\rm FDL}
  \label{eq:7}
 \end{aligned}
\end{equation}
where $\beta$, $\gamma$ and $\delta$ are the coefficients of the three losses. A point to note is that none of the target samples are involved in the pretraining on the source domain. In other words, the mini-batch $B^{\rm t}$ can be viewed as an empty set in the pretraining stage.

According to the relative research, the CE loss helps to cluster each ID class in the global scale, and the Triplet loss does well in correcting the boundaries between neighboring classes. Coupled with the proposed FDL loss, both the empirical error and the generalization error of the model can be properly balanced through the iterative training based on the overall loss given by \cref{eq:7}. Above all, the training algorithm of the proposed approach can be summarized as follows.

\begin{table}[!ht]
\begin{tabular*}{\hsize}{@{}@{\extracolsep{\fill}}l@{}}
\toprule
\textbf{Algorithm 1} Training process\\
\midrule
\textbf{Input:} Original source domain dataset $D^{\rm s}$, target domain dataset $D^{\rm t}$, the initialized model $\mathcal{M}$ , and \\
the Sample Dropout rate $\rho$ \\
\textbf{1:} Pretrain model $\mathcal{M}$ on the dataset $D^{\rm s}$ based on \cref{eq:7} \\
\textbf{2: For} $k=1,2,...,K$, \textbf{do}:\\
\textbf{3:}\qquad Randomly select a subset $D_k^{\rm t}$ from $D^{\rm t}$ for the clustering and training;\\
\textbf{4:}\qquad Extract the united features $\boldsymbol{\tilde{f}}$ for the dataset $D_k^{\rm t}$ with $\mathcal{M}$;\\
\textbf{5:}\qquad Assign pseudo labels for the dataset $D_k^{\rm t}$  based on their features;\\
\textbf{6:}\qquad Train $\mathcal{M}$ on the datasets $D^{\rm s}$ and $D_k^{\rm t}$ to minimize the overall loss in \cref{eq:7}\\
\textbf{7: Until} the model $\mathcal{M}$ converged\\
\textbf{Return} Final model $\mathcal{M}$ with optimal parameters\\
\bottomrule
\end{tabular*}
\end{table}

\section{Experiments}
\label{sec:exper}

\subsection{Datasets}
The proposed approach is evaluated on three popular datasets Market-1501\cite{ref:market1501}, DukeMTMC-reID\cite{ref:duke} and MSMT17\cite{ref:PTGAN}. The Market-1501 dataset contains 32,668 labeled images of 1501 identities from 6 disjoint cameras, in which the training set includes 751 identities and 12936 images, the gallery set contains 19732 images from 750 identities, and the query set contains 3368 images from 750 identities. The DukeMTMC-reID dataset includes 36411 images with 1402 identities, in which 16522 images of 702 identities are for training. The gallery and query sets contain 17,661 and 2228 images of another 702 identities. MSMT17 contains 126441 images of 4101 identities, of which 1041 identities and 3060 identities are used for training and testing respectively. 

\subsection{Settings}
\textbf{Implementation details:} The feature encoders $F_1$, $F_2$ and their mean encoders $\tilde{F}_1$ and $\tilde{F}_2$ adopt the ResNet-50\cite{ref:resnet} as backbone. They are initialized with the parameters pretrained on the ImageNet. Compared with the original ResNet-50, the fully connected layers of these encoders are discarded. The corresponding classifiers $C_1$ and $C_2$ will be trained directly with the given datasets. In the training process, each input image is uniformly resized to 256×128. Horizontal flipping, random cropping, and random erasing are performed to generate the augmented data. The mini-batch size is set as $|B^{\rm s}|=|B^{\rm t}|=60$, in which each identity contains 4 different samples. Hyperparameters $\alpha$, $\beta$, $\gamma$, $\delta$ and $\tau$ in \cref{eq:1}, \cref{eq:7} and \cref{eq:5} are set as 0.999, 1, 1, 0.5 and 0.3 respectively. The training is implemented by the Adam optimizer with the learning rate schedule where $\eta=0.00035$ at beginning and is divided by 10 after every 20 epochs. The whole training process is finished after 55 epochs, in which the first epoch is used to pretrain the model with only the source domain dataset and the other epochs are used to train the model with both the source and target datasets.

\textbf{Evaluation metrics:} In the testing process, cumulative matching characteristics at Rank-1, Rank-5, Rank-10 and mAP are applied to evaluate the performance of our method.

\subsection{Comparison with the State-of-the-art Methods}
We compare our proposed FDL-SD method with other SOTA unsupervised domain adaptive ReID works on the Market-1501, DukeMTMC-reID and MSMT17 datasets. The comparison results are shown in \cref{tab:sota}. 

\begin{table*}
    \centering
    \begin{tabular}{cccccccccc}
        \toprule
        \multirow{2}{*}{Methods} & \multicolumn{4}{c}{Market to Duke} & \multicolumn{1}{c}{} & \multicolumn{4}{c}{Duke to Market}\\
        \cline{2-5}
        \cline{7-10}
        & mAP & R1 & R5 & R10 & & mAP & R1 & R5 & R10\\
        \midrule
        PTGAN\cite{ref:PTGAN}(CVPR’18)	&\_	&27.4	&\_	&50.7	& &\_	&38.6	&\_	&66.1\\
        SPGAN\cite{ref:SPGAN}(CVPR’18)	&22.3	&41.1	&56.6	&63.0	& &22.8	&51.5	&70.1	&76.8\\
        ECN\cite{ref:ECN}(CVPR’19)	&40.4	&63.3	&75.8	&80.4	& &43.0	&75.1	&87.6	&91.6\\
        CR-GAN\cite{ref:CRGAN}(ICCV’19) &48.6	&68.9	&80.2	&84.7	& &54.0	&77.7	&89.7	&92.7\\
        SSG\cite{ref:SSG}(ICCV’19)	&53.4	&73.0	&80.6	&83.2	& &58.3	&80.0	&90.0	&92.4\\
        D-MMD\cite{ref:DMMD}(CVPR’20)	    &46.0	&63.5	&78.8	&83.9	& &48.8	&70.6	&87.0	&91.5\\
        DAAM\cite{ref:DAAM}(AAAI’20)	    &48.8	&71.3	&82.4	&86.3	& &53.1	&77.8	&89.9	&93.7\\
        SADA\cite{ref:SADA}(CVPR’20)	&55.8	&74.5	&85.3	&88.7	& &59.8	&83.0	&91.8	&94.1\\
        NRMT\cite{ref:NRMT}(ECCV’20)	&62.2	&77.8	&86.9	&89.5	& &71.7	&87.8	&94.6	&96.5\\
        MMT\cite{ref:MMT}(ICLR’20)	&65.1	&78.0	&88.8	&92.5	& &71.2	&87.7	&94.9	&96.9\\
        SpCL\cite{ref:SpCL}(NeurIPS’20)	&68.8	&82.9	&90.1	&92.5	& &76.7	&90.3	&96.2	&97.7\\
        MMT+RDSBN\cite{ref:MMT+}(CVPR’21)	&66.6	&80.3	&89.1	&92.6	& &81.5	&92.9	&97.6	&98.4\\
        UNRN\cite{ref:UNRN}(AAAI’21)	    &69.1	&82.0	&90.7	&93.5	& &78.1	&91.9	&96.1	&97.8\\
        RLCC\cite{ref:RLCC}(CVPR’21)	    &69.2	&83.2	&91.6	&93.8	& &77.7	&90.8	&96.3	&97.5\\
        GLT\cite{ref:GLT}(CVPR’21)	    &69.2	&82.0	&90.2	&92.8	& &79.5	&92.2	&96.5	&97.8\\
        \cline{1-10}
        \textbf{Ours(FDL-SD)} &\textbf{71.3}	&\textbf{83.6}	&\textbf{91.2}	&\textbf{93.4}	& &\textbf{83.0}	&\textbf{93.2}	&\textbf{97.6}	&\textbf{98.3}\\
        \hline
        \hline
       
        \multirow{2}{*}{Methods} & \multicolumn{4}{c}{Market to MSMT} & \multicolumn{1}{c}{} & \multicolumn{4}{c}{Duke to MSMT}\\
       \cline{2-5}
        \cline{7-10}
        & mAP & R1 & R5 & R10 & & mAP & R1 & R5 & R10\\
        \midrule
        ECN\cite{ref:ECN}(CVPR’19)	&8.5	&25.3	&36.3	&42.1	& &10.2	&30.2	&41.5	&46.8\\
        SSG\cite{ref:SSG}(ICCV’19)	&13.2	&31.6	&-	&49.6	& &13.3	&32.2	&-	&51.2\\
        DAAM\cite{ref:DAAM}(AAAI’20)	&20.8	&44.5	&-	&-	& &21.6	&46.7	&-	&-\\
        NRMT\cite{ref:NRMT}(ECCV’20)	&19.8	&43.7	&56.5   &62.2	& &20.6	&45.2	&57.8	&63.3\\
        MMT\cite{ref:MMT}(ICLR’20)	&22.9	&49.2	&63.1	&68.8	& &23.3	&50.1	&63.9	&69.8\\
        RDSBN\cite{ref:MMT+}(CVPR’21)	&20.7	&46.9	&60.0	&65.0	& &21.3	&47.7	&60.8	&66.4\\
        UNRN\cite{ref:UNRN}(AAAI’21)	&25.3	&52.4	&64.7	&69.7	& &26.2	&54.9	&67.3	&70.6\\
        GLT\cite{ref:GLT}(CVPR’21)	&26.5	&56.6	&67.5	&72.0	& &27.7	&59.5	&70.1	&74.2\\

        \cline{1-10}
        \textbf{Ours(FDL-SD)} &\textbf{26.6}	&\textbf{53.7}	&\textbf{67.0}	&\textbf{72.3}	& &\textbf{31.2}	&\textbf{60.1}	&\textbf{72.6}	&\textbf{77.4}\\
        \bottomrule
    \end{tabular}
    \caption{Performance (\%) comparison with some popular SOTA unsupervised domain adaptive ReID methods on Market-1501, DukeMTMC-reID and MSMT17.}
    \label{tab:sota}
\end{table*}

At first, we compare the ReID performance on Market-1501 and DukeMTMC-reID. Some typical works along the approach of domain bias reduction, including PTGAN\cite{ref:PTGAN}, SPGAN\cite{ref:SPGAN} and CR-GAN\cite{ref:CRGAN} which are GAN-based, DAAM\cite{ref:DAAM}, D-MMD\cite{ref:DMMD} and SADA\cite{ref:SADA} which are based on feature alignment are compared with our approach. The results show that our model significantly outperforms the best of these models by about 15-23\% (mAP: 71.3\% vs 55.8\% on the Market-to-Duke task, and 83.0\% vs 59.8\% on the Duke-to-Market task). Comparing the traditional clustering-based methods like SSG\cite{ref:SSG} and ECN\cite{ref:ECN}, our model is still more superior in all the metrics on both the adaptation tasks. There are also a lot of latest models tackling the noisy problem such as NRMT\cite{ref:NRMT}, SpCL\cite{ref:SpCL}, UNRN\cite{ref:UNRN} and GLT\cite{ref:GLT} in the comparison. Even these methods have achieved better performance in comparison with the above approaches, our method can still beat them on all the evaluation scores. Especially, our method outperforms GLT\cite{ref:GLT} by 2.1\%/1.6\% on mAP/Rank-1 scores for the Market-to-Duke task, and also takes an advantage of 3.5\%/1\% on the Duke-to-Market task. For the most challenging tasks of Market-to-MSMT and Duke-to-MSMT, our approach also has significant improvement, especially for the mAP score on Duke-to-MSMT task, our work outperforms the existing works by a large margin. These results strongly support the superiority of our method over the SOTA competing approaches.

\subsection{Analysis for Sample Dropout Rate $\rho$}
It has been found from the experimental results that the Sample Dropout rate $\rho$ has a significant impact on the final performance of the model. To clarify this influence, a group of tests are carried out to analyse the underlying relationship between the parameter $\rho$, the ReID metrics and the clustering results.

\textbf{The influence on performance metrics:} We have uniformly sampled the values of $\rho$ from 0 to 0.8. The metric scores under different values of $\rho$ are shown in \cref{tab:SD}. It can be observed that the performance on all the tasks has been boosted clearly as the parameter $\rho$ increases from 0. It reaches the peak when $\rho=0.4$ and falls back when $\rho$ continues increasing. The results prove that the proposed SD method plays a notable role to affect the training results, and a proper value of $\rho$ can significantly improve the performance of the model.

\begin{table*}
    \centering
    \begin{tabular}{ccccccccccccc}
        \toprule
        \multirow{2}{*}{Value of  $\rho$} & \multicolumn{2}{c}{Market to Duke} & \multicolumn{1}{c}{} & \multicolumn{2}{c}{Duke to Market} & \multicolumn{1}{c}{} & \multicolumn{2}{c}{Market to MSMT} & \multicolumn{1}{c}{} & \multicolumn{2}{c}{Duke to MSMT}  \\
        \cline{2-3}
        \cline{5-6}
        \cline{8-9}
        \cline{11-12}
        
        & mAP & R1 & & mAP & R1 & & mAP & R1 & &mAP & R1\\
        \midrule
        0	&65.3	&79.5	 & &74.1	&89.8  & &22.4 &49.0  & &26.7 &55.5\\
        0.1	&67.4	&80.7	 & &77.4	&90.5  & &23.7 &50.9  & &27.8 &56.7\\
        0.2	&69.4	&83.3	 & &81.1	&92.7  & &25.2 &52.2  & &29.2 &57.8\\
        0.3	&70.8	&83.4    & &80.4	&92.1  & &24.1 &50.9  & &29.6 &58.6\\
        \textcolor{red}{0.4}	&\textcolor{red}{71.3}	&\textcolor{red}{83.6}	& &\textcolor{red}{83.0}	&\textcolor{red}{93.2}  & &\textcolor{red}{26.6} &\textcolor{red}{53.7} & &\textcolor{red}{31.2} &\textcolor{red}{60.1}\\
        0.5	&71.0	&83.1	 & &81.1	&92.3   & &25.3 &51.6  & &29.2 &57.7\\
        0.6	&70.0	&83.0	 & &79.7	&91.7	& &19.7	&42.7  & &28.1	&55.7\\
        0.7	&65.9	&79.7	 & &75.1	&88.8	& &16.1	&36.3  & &22.6	&46.9\\
        0.8	&52.8	&70.4	 & &47.3	&71.3	& &5.5	&13.6  & &12.5	&30.0\\
        \bottomrule
    \end{tabular}
    \caption{The performance (\%) of the model on four tasks when different values of $\rho$ are applied.}
    \label{tab:SD}
\end{table*}

\begin{figure*}
  \centering
  \includegraphics[width=0.99\textwidth]{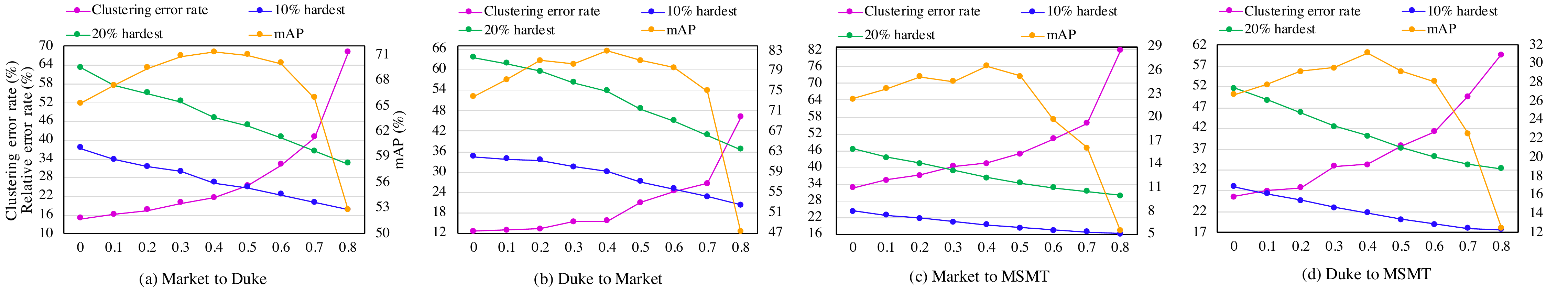}
  \caption{Curves of clustering error rate, relative error rate of the 10\% and 20\% hardest samples and the mAP plotted along with growing values of the parameter $\rho$. Similar to (a), the left vertical axes of (b), (c) and (d) represent the clustering error rate and relative error rate, the right axes indicate the mAP scores.}
  \label{fig:cluster}
\end{figure*}

\textbf{The influence on clustering result:} To deeply dig out the underlying mechanism of the proposed SD method, we design a group of experiments to observe the distributions of noisy pseudo labels for different values of $\rho$. According to the proposed clustering-based method, a target sample $\textbf{\emph{x}}_i^{\rm t}$ in the $k$-th clustering and training process will be assigned with a noisy pseudo label when its true label does not equal the dominant true label of the corresponding cluster. The more times the noisy pseudo labels are assigned in the iterative epochs for a sample, the harder the sample is. According to the results, the curves of four metrics against the increasing values of $\rho$ are plotted in \cref{fig:cluster}. 

The clustering error rate is the ratio of noisy labels to all the labels when we implement the clustering algorithm on the fully converged model. The corresponding pink curves in \cref{fig:cluster}(a), (b), (c) and (d) illustrate that the clustering error rate increases with the growing value of $\rho$. This result coincides with our intuition because higher SD rate indicates that more samples are not fully utlized in the training. The blue and green curves in \cref{fig:cluster} record the relative error rates of the 10\% and 20\% hardest samples against the parameter $\rho$, respectively. By 10\% or 20\% hardest, we mean the top 10\% or 20\% samples that are most frequently assigned with noisy pseudo labels. The corresponding relative error rates are defined as the ratio of the noisy labels ever assigned to these hardest samples to all the noisy labels ever generated in the whole training process. It can be easily discovered that the relative error rates of the 10\% or 20\% hardest samples fall steadily as the value of $\rho$ grows from 0 to 0.8. This result reveals a truth that the proposed SD method can effectively prevent noisy pseudo labels from concentrating on a minority of target samples, namely the hardest samples we mentioned above. 

\begin{figure*}
  \centering
  \includegraphics[width=0.99\textwidth]{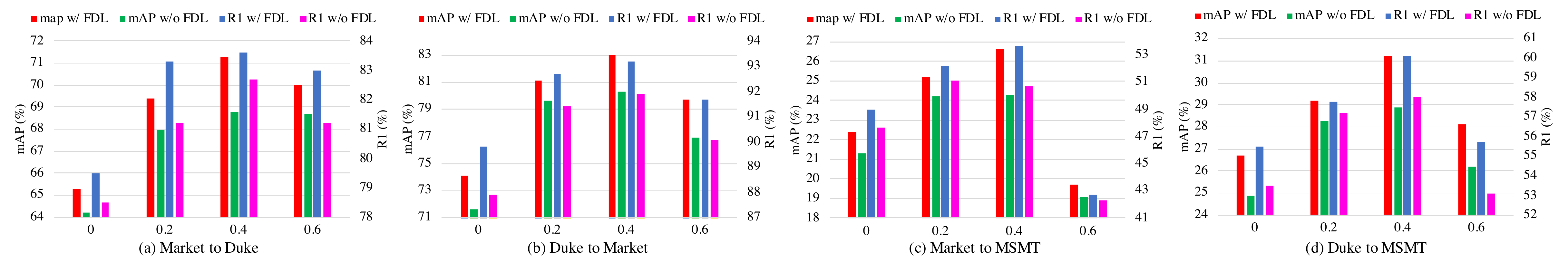}
  \caption{The mAP and Rank-1 scores for ablation tests under different values of $\rho$ on four tasks, where “w/” means the FDL is applied to the model while “w/o” means the opposite.}
  \label{fig:FDL}
\end{figure*}

Considering the synthetic influence of the hyperparameter $\rho$ on both aspects of the clustering error rate of noisy labels and the relative error rate of the hardest samples, a clear conclusion can be drawn that there must be a most proper value of $\rho$ that can keep the optimal balance between the two factors. The mAP curve in \cref{fig:cluster} also provides solid evidence to support this point, in which the mAP score reaches the peak at $\rho=0.4$ for the four tasks. All above experimental results and the corresponding analysis prove that the proposed SD method can significantly suppress the ill influence caused by hard samples and consequently improve the reliability of the model.

\subsection{Ablation Study on FDL}
To validate the effectiveness of our proposed Feature Diversity Learning, a group of ablation tests are designed to show the mAP and Rank-1 scores of the model when the FDL is or is not applied. As shown as \cref{fig:FDL}, against different values of $\rho$ in the horizontal axes, the red and green bars represent the mAP scores with or without the FDL, and the blue and pink bars reflect the Rank-1 scores when the FDL is or is not applied, respectively. It can be clearly seen that the FDL achieves 2.8\% and 1.2\% improvements on mAP and Rank-1 scores in dealing with the Market-to-Duke task when $\rho=0.4$, and it also shows 2.7\% and 1.4\% superiority of mAP and Rank-1 scores on the Duke-to-Market task. This advantage can also be validated on both of the Market-to-MSMT and Duke-to-MSMT tasks. The results under different values of $\rho$ on 
all the tasks indicate the same conclusion that the proposed FDL certainly improves the generalization ability of the model on the unsupervised domain adaptive ReID task, which supports our viewpoint that the diversity between the two feature streams helps to prevent the model from falling into the overfitting trap.

\section{Conclusion}
Noisy pseudo label is one of the most challenging problem for clustering-based unsupervised domain adaptive ReID models, which is the main issue we have addressed in this paper. With our proposed Sample Dropout method, the hard samples can be effectively suppressed and consequently the uneven distribution of noisy labels can be smoothed, which is proved helpful for breaking the vicious circle between noisy label and bad feature and improving the training results. The proposed Feature Diversity Learning provides a new approach for the well-known mutual teaching architecture, which focuses on enhancing the diversity of the two feature streams. Ablation study shows that FDL has a stable positive impact on the generalization ability of the model. With the above two improvements, it is proved by the comparison results that our proposed FDL-SD outperforms most state-of-the-art methods on the unsupervised domain adaptive ReID task.

\bibliographystyle{plain}
\bibliography{sample}

\begin{thebibliography}{10}

\bibitem{ref:DBSCAN}
Henrik B{\"a}cklund, Anders Hedblom, and Niklas Neijman.
\newblock A density-based spatial clustering of application with noise.
\newblock {\em Data Mining TNM033}, pages 11--30, 2011.

\bibitem{ref:MMT+}
Zechen Bai, Zhigang Wang, Jian Wang, Di~Hu, and Errui Ding.
\newblock Unsupervised multi-source domain adaptation for person
  re-identification.
\newblock In {\em Proceedings of the IEEE/CVF Conference on Computer Vision and
  Pattern Recognition}, pages 12914--12923, 2021.

\bibitem{ref:DCML}
Guangyi Chen, Yuhao Lu, Jiwen Lu, and Jie Zhou.
\newblock Deep credible metric learning for unsupervised domain adaptation
  person re-identification.
\newblock In {\em Computer Vision--ECCV 2020: 16th European Conference,
  Glasgow, UK, August 23--28, 2020, Proceedings, Part VIII 16}, pages 643--659.
  Springer, 2020.

\bibitem{ref:CL3}
Ting Chen, Simon Kornblith, Mohammad Norouzi, and Geoffrey Hinton.
\newblock A simple framework for contrastive learning of visual
  representations.
\newblock In {\em International conference on machine learning}, pages
  1597--1607. PMLR, 2020.

\bibitem{ref:CRGAN}
Yanbei Chen, Xiatian Zhu, and Shaogang Gong.
\newblock Instance-guided context rendering for cross-domain person
  re-identification.
\newblock In {\em Proceedings of the IEEE/CVF International Conference on
  Computer Vision}, pages 232--242, 2019.

\bibitem{ref:learningwith}
Hao Cheng, Zhaowei Zhu, Xingyu Li, Yifei Gong, Xing Sun, and Yang Liu.
\newblock Learning with instance-dependent label noise: A sample sieve
  approach.
\newblock {\em arXiv preprint arXiv:2010.02347}, 2020.

\bibitem{ref:SPGAN}
Weijian Deng, Liang Zheng, Qixiang Ye, Guoliang Kang, Yi~Yang, and Jianbin
  Jiao.
\newblock Image-image domain adaptation with preserved self-similarity and
  domain-dissimilarity for person re-identification.
\newblock In {\em Proceedings of the IEEE conference on computer vision and
  pattern recognition}, pages 994--1003, 2018.

\bibitem{ref:SSG}
Yang Fu, Yunchao Wei, Guanshuo Wang, Yuqian Zhou, Honghui Shi, and Thomas~S
  Huang.
\newblock Self-similarity grouping: A simple unsupervised cross domain
  adaptation approach for person re-identification.
\newblock In {\em Proceedings of the IEEE/CVF International Conference on
  Computer Vision}, pages 6112--6121, 2019.

\bibitem{ref:MMT}
Yixiao Ge, Dapeng Chen, and Hongsheng Li.
\newblock Mutual mean-teaching: Pseudo label refinery for unsupervised domain
  adaptation on person re-identification.
\newblock In {\em International Conference on Learning Representations}, 2020.

\bibitem{ref:SpCL}
Yixiao Ge, Feng Zhu, Dapeng Chen, Rui Zhao, and Hongsheng Li.
\newblock Self-paced contrastive learning with hybrid memory for domain
  adaptive object re-id.
\newblock In {\em Advances in Neural Information Processing Systems}, 2020.

\bibitem{ref:MAE}
Aritra Ghosh, Himanshu Kumar, and PS~Sastry.
\newblock Robust loss functions under label noise for deep neural networks.
\newblock In {\em Proceedings of the AAAI Conference on Artificial
  Intelligence}, volume~31, 2017.

\bibitem{ref:Co}
Bo~Han, Quanming Yao, Xingrui Yu, Gang Niu, Miao Xu, Weihua Hu, Ivor Tsang, and
  Masashi Sugiyama.
\newblock Co-teaching: Robust training of deep neural networks with extremely
  noisy labels.
\newblock {\em arXiv preprint arXiv:1804.06872}, 2018.

\bibitem{ref:CL2}
Kaiming He, Haoqi Fan, Yuxin Wu, Saining Xie, and Ross Girshick.
\newblock Momentum contrast for unsupervised visual representation learning.
\newblock In {\em Proceedings of the IEEE/CVF Conference on Computer Vision and
  Pattern Recognition}, pages 9729--9738, 2020.

\bibitem{ref:resnet}
Kaiming He, Xiangyu Zhang, Shaoqing Ren, and Jian Sun.
\newblock Deep residual learning for image recognition.
\newblock In {\em Proceedings of the IEEE conference on computer vision and
  pattern recognition}, pages 770--778, 2016.

\bibitem{ref:EaNet}
Houjing Huang, Wenjie Yang, Xiaotang Chen, Xin Zhao, Kaiqi Huang, Jinbin Lin,
  Guan Huang, and Dalong Du.
\newblock Eanet: Enhancing alignment for cross-domain person re-identification.
\newblock {\em arXiv preprint arXiv:1812.11369}, 2018.

\bibitem{ref:DAAM}
Yangru Huang, Peixi Peng, Yidong Li, Yi~Jin, Junliang Xing, and Shiming Ge.
\newblock Domain adaptive attention model for unsupervised cross-domain person
  re-identification.
\newblock AAAI, 2020.

\bibitem{ref:twostagecluster}
Zilong Ji, Xiaolong Zou, Xiaohan Lin, Xiao Liu, Tiejun Huang, and Si~Wu.
\newblock An attention-driven two-stage clustering method for unsupervised
  person re-identification.
\newblock In {\em Computer Vision--ECCV 2020: 16th European Conference,
  Glasgow, UK, August 23--28, 2020, Proceedings, Part XXVIII 16}, pages 20--36.
  Springer, 2020.

\bibitem{ref:cleannet}
Kuang-Huei Lee, Xiaodong He, Lei Zhang, and Linjun Yang.
\newblock Cleannet: Transfer learning for scalable image classifier training
  with label noise.
\newblock In {\em Proceedings of the IEEE Conference on Computer Vision and
  Pattern Recognition}, pages 5447--5456, 2018.

\bibitem{ref:JVTC}
Jianing Li and Shiliang Zhang.
\newblock Joint visual and temporal consistency for unsupervised domain
  adaptive person re-identification.
\newblock In {\em European Conference on Computer Vision}, pages 483--499.
  Springer, 2020.

\bibitem{ref1:1}
Jianing Li, Shiliang Zhang, Qi~Tian, Meng Wang, and Wen Gao.
\newblock Pose-guided representation learning for person re-identification.
\newblock {\em IEEE transactions on pattern analysis and machine intelligence},
  2019.

\bibitem{ref1:3}
Wei Li, Rui Zhao, Tong Xiao, and Xiaogang Wang.
\newblock Deepreid: Deep filter pairing neural network for person
  re-identification.
\newblock In {\em Proceedings of the IEEE conference on computer vision and
  pattern recognition}, pages 152--159, 2014.

\bibitem{ref:BUG}
Yutian Lin, Xuanyi Dong, Liang Zheng, Yan Yan, and Yi~Yang.
\newblock A bottom-up clustering approach to unsupervised person
  re-identification.
\newblock In {\em Proceedings of the AAAI Conference on Artificial
  Intelligence}, volume~33, pages 8738--8745, 2019.

\bibitem{ref1:7}
Jiawei Liu, Zheng-Jun Zha, Di~Chen, Richang Hong, and Meng Wang.
\newblock Adaptive transfer network for cross-domain person re-identification.
\newblock In {\em Proceedings of the IEEE/CVF Conference on Computer Vision and
  Pattern Recognition}, pages 7202--7211, 2019.

\bibitem{ref:DMMD}
Djebril Mekhazni, Amran Bhuiyan, George Ekladious, and Eric Granger.
\newblock Unsupervised domain adaptation in the dissimilarity space for person
  re-identification.
\newblock In {\em European Conference on Computer Vision}, pages 159--174.
  Springer, 2020.

\bibitem{ref:duke}
Ergys Ristani, Francesco Solera, Roger Zou, Rita Cucchiara, and Carlo Tomasi.
\newblock Performance measures and a data set for multi-target, multi-camera
  tracking.
\newblock In {\em European conference on computer vision}, pages 17--35.
  Springer, 2016.

\bibitem{ref2:1}
L.~Shan, H.~Li, C.~T. Li, and A.~C. Kot.
\newblock Multi-task mid-level feature alignment network for unsupervised
  cross-dataset person re-identification.
\newblock In {\em BMVC 2018}, 2018.

\bibitem{ref:meta-weight}
Jun Shu, Qi~Xie, Lixuan Yi, Qian Zhao, Sanping Zhou, Zongben Xu, and Deyu Meng.
\newblock Meta-weight-net: Learning an explicit mapping for sample weighting.
\newblock {\em arXiv preprint arXiv:1902.07379}, 2019.

\bibitem{ref1:2}
Chi Su, Jianing Li, Shiliang Zhang, Junliang Xing, Wen Gao, and Qi~Tian.
\newblock Pose-driven deep convolutional model for person re-identification.
\newblock In {\em Proceedings of the IEEE international conference on computer
  vision}, pages 3960--3969, 2017.

\bibitem{ref1:4}
Yifan Sun, Liang Zheng, Yi~Yang, Qi~Tian, and Shengjin Wang.
\newblock Beyond part models: Person retrieval with refined part pooling (and a
  strong convolutional baseline).
\newblock pages 480--496, 2018.

\bibitem{ref1:5}
Chiat-Pin Tay, Sharmili Roy, and Kim-Hui Yap.
\newblock Aanet: Attribute attention network for person re-identifications.
\newblock In {\em Proceedings of the IEEE/CVF Conference on Computer Vision and
  Pattern Recognition}, pages 7134--7143, 2019.

\bibitem{ref:towards}
Arash Vahdat.
\newblock Toward robustness against label noise in training deep discriminative
  neural networks.
\newblock {\em arXiv preprint arXiv:1706.00038}, 2017.

\bibitem{ref:MMCL}
Dongkai Wang and Shiliang Zhang.
\newblock Unsupervised person re-identification via multi-label classification.
\newblock In {\em Proceedings of the IEEE/CVF Conference on Computer Vision and
  Pattern Recognition}, pages 10981--10990, 2020.

\bibitem{ref:SADA}
Guangcong Wang, Jian-Huang Lai, Wenqi Liang, and Guangrun Wang.
\newblock Smoothing adversarial domain attack and p-memory reconsolidation for
  cross-domain person re-identification.
\newblock In {\em Proceedings of the IEEE/CVF Conference on Computer Vision and
  Pattern Recognition}, pages 10568--10577, 2020.

\bibitem{ref:SCE}
Yisen Wang, Xingjun Ma, Zaiyi Chen, Yuan Luo, Jinfeng Yi, and James Bailey.
\newblock Symmetric cross entropy for robust learning with noisy labels.
\newblock In {\em Proceedings of the IEEE/CVF International Conference on
  Computer Vision}, pages 322--330, 2019.

\bibitem{ref:PTGAN}
Longhui Wei, Shiliang Zhang, Wen Gao, and Qi~Tian.
\newblock Person transfer gan to bridge domain gap for person
  re-identification.
\newblock In {\em Proceedings of the IEEE conference on computer vision and
  pattern recognition}, pages 79--88, 2018.

\bibitem{ref:CL1}
Zhirong Wu, Yuanjun Xiong, Stella~X Yu, and Dahua Lin.
\newblock Unsupervised feature learning via non-parametric instance
  discrimination.
\newblock In {\em Proceedings of the IEEE conference on computer vision and
  pattern recognition}, pages 3733--3742, 2018.

\bibitem{ref:OIM}
Tong Xiao, Shuang Li, Bochao Wang, Liang Lin, and Xiaogang Wang.
\newblock Joint detection and identification feature learning for person
  search.
\newblock In {\em Proceedings of the IEEE Conference on Computer Vision and
  Pattern Recognition}, pages 3415--3424, 2017.

\bibitem{ref1:6}
Hong-Xing Yu, Wei-Shi Zheng, Ancong Wu, Xiaowei Guo, Shaogang Gong, and
  Jian-Huang Lai.
\newblock Unsupervised person re-identification by soft multilabel learning.
\newblock In {\em Proceedings of the IEEE/CVF Conference on Computer Vision and
  Pattern Recognition}, pages 2148--2157, 2019.

\bibitem{ref:HCT}
Kaiwei Zeng, Munan Ning, Yaohua Wang, and Yang Guo.
\newblock Hierarchical clustering with hard-batch triplet loss for person
  re-identification.
\newblock In {\em Proceedings of the IEEE/CVF Conference on Computer Vision and
  Pattern Recognition}, pages 13657--13665, 2020.

\bibitem{ref:MEBNet}
Yunpeng Zhai, Qixiang Ye, Shijian Lu, Mengxi Jia, Rongrong Ji, and Yonghong
  Tian.
\newblock Multiple expert brainstorming for domain adaptive person
  re-identification.
\newblock In {\em Computer Vision--ECCV 2020: 16th European Conference,
  Glasgow, UK, August 23--28, 2020, Proceedings, Part VII 16}, pages 594--611.
  Springer, 2020.

\bibitem{ref:RLCC}
Xiao Zhang, Yixiao Ge, Yu~Qiao, and Hongsheng Li.
\newblock Refining pseudo labels with clustering consensus over generations for
  unsupervised object re-identification.
\newblock In {\em Proceedings of the IEEE/CVF Conference on Computer Vision and
  Pattern Recognition}, pages 3436--3445, 2021.

\bibitem{ref:GCE}
Zhilu Zhang and Mert~R Sabuncu.
\newblock Generalized cross entropy loss for training deep neural networks with
  noisy labels.
\newblock In {\em 32nd Conference on Neural Information Processing Systems
  (NeurIPS)}, 2018.

\bibitem{ref:NRMT}
Fang Zhao, Shengcai Liao, Guo-Sen Xie, Jian Zhao, Kaihao Zhang, and Ling Shao.
\newblock Unsupervised domain adaptation with noise resistible mutual-training
  for person re-identification.
\newblock In {\em European Conference on Computer Vision}, pages 526--544.
  Springer, 2020.

\bibitem{ref:UNRN}
Kecheng Zheng, Cuiling Lan, Wenjun Zeng, Zhizheng Zhang, and Zheng-Jun Zha.
\newblock Exploiting sample uncertainty for domain adaptive person
  re-identification.
\newblock In {\em AAAI 2021}. AAAI, February 2021.

\bibitem{ref:GLT}
Kecheng Zheng, Wu~Liu, Lingxiao He, Tao Mei, Jiebo Luo, and Zheng-Jun Zha.
\newblock Group-aware label transfer for domain adaptive person
  re-identification.
\newblock In {\em Proceedings of the IEEE/CVF Conference on Computer Vision and
  Pattern Recognition}, pages 5310--5319, 2021.

\bibitem{ref:market1501}
Liang Zheng, Liyue Shen, Lu~Tian, Shengjin Wang, Jingdong Wang, and Qi~Tian.
\newblock Scalable person re-identification: A benchmark.
\newblock In {\em Proceedings of the IEEE international conference on computer
  vision}, pages 1116--1124, 2015.

\bibitem{ref:ECN}
Zhun Zhong, Liang Zheng, Zhiming Luo, Shaozi Li, and Yi~Yang.
\newblock Invariance matters: Exemplar memory for domain adaptive person
  re-identification.
\newblock In {\em Proceedings of the IEEE/CVF Conference on Computer Vision and
  Pattern Recognition}, pages 598--607, 2019.

\bibitem{ref:DGNET}
Yang Zou, Xiaodong Yang, Zhiding Yu, BVK~Vijaya Kumar, and Jan Kautz.
\newblock Joint disentangling and adaptation for cross-domain person
  re-identification.
\newblock In {\em Computer Vision--ECCV 2020: 16th European Conference,
  Glasgow, UK, August 23--28, 2020, Proceedings, Part II 16}, pages 87--104.
  Springer, 2020.

\end{thebibliography}
\end{document}